\documentclass[10pt,twocolumn,letterpaper]{article}

\usepackage[pagenumbers]{cvpr} 
\usepackage{makecell}
\usepackage{multirow}
\usepackage{subcaption}
\usepackage{booktabs}
\usepackage{diagbox}
\usepackage{colortbl}
\usepackage{bbding}
\definecolor{mygray}{gray}{.9}

\usepackage[dvipsnames]{xcolor}

\definecolor{cvprblue}{rgb}{0.21,0.49,0.74}
\usepackage[pagebackref,breaklinks,colorlinks,citecolor=cvprblue]{hyperref}

\def\paperID{1827} 
\def\confName{CVPR}
\def\confYear{2024}

\title{Decoupled DETR For Few-shot Object Detection}

\author{Zeyu Shangguan, Lian Huai, Tong Liu, Xingqun Jiang\\
BOE Technology Group Co., Ltd.\\
No.09 Dize Road, Daxing District, Beijing, China\\
{\tt\small \{shangguanzeyu,huailian,liutongcto,jiangxingqun\}@boe.com.cn}
}

\begin{document}
\maketitle

\begin{abstract}
Few-shot object detection (FSOD), an efficient method for addressing the severe data-hungry problem, has been extensively discussed.
Current works have significantly advanced the problem in terms of model and data.
However, the overall performance of most FSOD methods still does not fulfill the desired accuracy.
In this paper we improve the FSOD model to address the severe issue of sample imbalance and weak feature propagation.
To alleviate modeling bias from data-sufficient base classes, we examine the effect of decoupling the parameters for classes with sufficient data and classes with few samples in various ways. We design a base-novel categories \textbf{de}coupled \textbf{DETR} (DeDETR) for FSOD.
We also explore various types of skip connection between the encoder and decoder for DETR.
Besides, we notice that the best outputs could come from the intermediate layer of the decoder instead of the last layer; therefore, we build a unified decoder module that could dynamically fuse the decoder layers as the output feature.
We evaluate our model on commonly used datasets such as PASCAL VOC and MSCOCO. Our results indicate that our proposed module could achieve stable improvements of 5\% to 10\% in both fine-tuning and meta-learning paradigms and has outperformed the highest score in recent works.
\end{abstract}
  
\section{Introduction}
\label{sec:intro}

Few-shot learning (FSL) aims to training a generalized deep learning model that can address the issue when training samples are extremely insufficient and often unseen before. This process mimics human infant learning, as infants can quickly learn new knowledge with very little instruction based on abundant already-learned knowledge. Few-shot learning has broad and promising applications in practice, such as industrial defect detection, medical image analysis, archaeological research, landform change detection, \etc.

\begin{figure}
	\begin{subfigure}{0.5\linewidth}
		\centering
		\includegraphics[width=40mm]{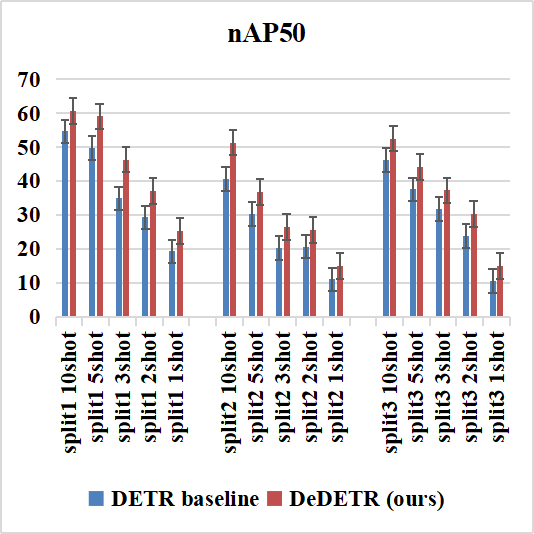}
		\caption{fine-tune}
		\label{fig:intro_finetune}
	\end{subfigure}
	\begin{subfigure}{0.45\linewidth}
		\centering
		\includegraphics[width=40mm]{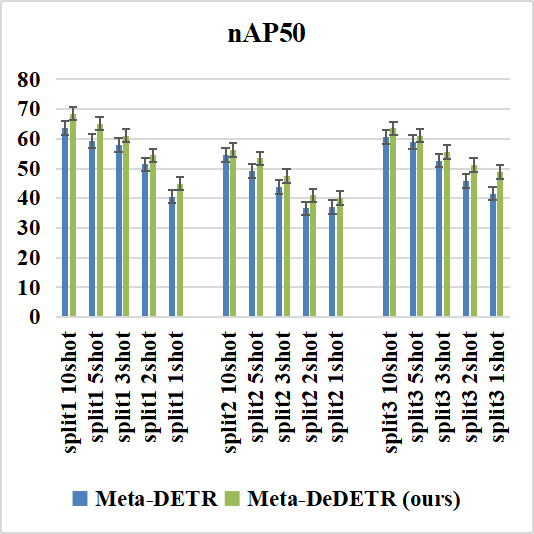}
		\caption{meta-learning}
		\label{fig:intro_metalrn}
	\end{subfigure}
	\caption{Our decoupled DETR achieves stable improvements under both fine-tune and meta-learning paradigms. Histograms demonstrate our experimental results for average precision of novel categories (nAP50) on the three few-shot PASCAL VOC data splits.}
	\label{fig:intro}
\end{figure}

\begin{figure*}[t]
  \centering
   \includegraphics[width=0.9\linewidth]{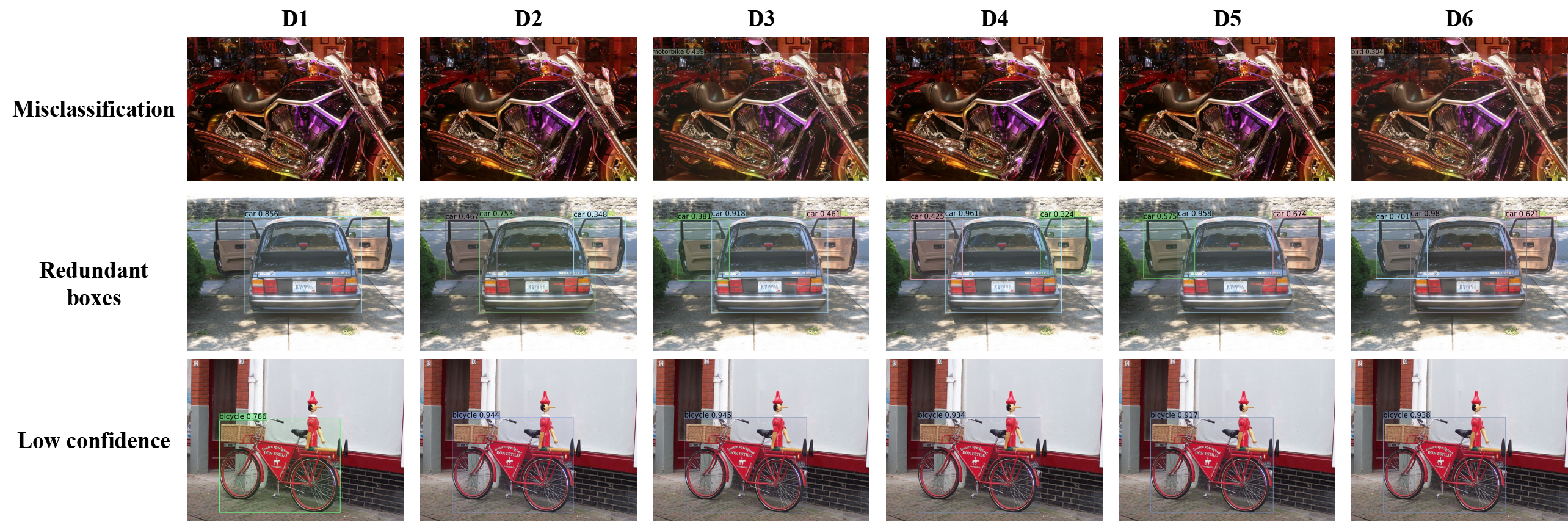}
   \caption{The intermediate layer of decoder may have better outputs comparing to the last layer of decoder, in terms of classification, boxes regression and prediction confidence.}
   \label{fig:decoder_layers}
\end{figure*}

Few-shot object detection(FSOD) is an important task in few-shot learning and holds practical significance in various scenarios. Since TFA~\cite{Wang20TFA}, FSOD has made significant progress based on the Faster RCNN (FRCN)~\cite{Ren15FRCN} baseline. There are two commonly used training paradigms, namely, meta-learning and fine-tuning. Current methods utilizing FRCN in meta-learning and fine-tuning training paradigms have achieved competitive results. The emergence of the detection transformer (DETR)~\cite{Carion20DETR} in 2020 has further improved the framework for general object detection. This end-to-end set prediction-based object detection framework has not only outperformed traditional anchor-based methods (FRCN, YOLO~\cite{Bochkovskiy20YOLOv4}, etc.) but has also been widely applied to more sub-tasks of object detection, including instance segmentation and FSOD.

As a result, FSOD based on DETR is considered as a new trend, not only due to the simplicity of the DETR framework, but also because of its homology to Transformer, which makes it easier to combine with other Transformer-based tasks, especially for multi-modal language-vision tasks~\cite{han2023multimodal}. Meanwhile, recent literature has demonstrated that DETR has achieved outstanding performance in FSOD. Meta-DETR~\cite{Zhang23MetaDETR} was the first to explore FSOD based on the meta-learning paradigm, while FsDETR~\cite{Bulat23FSDETR} was the first to explore FSOD without retraining. However, the development of FSOD based on DETR is still in its early stages. Currently, there is no baseline for fine-tuning, and we are the first to explore this paradigm.

FSL has been previously described as an extreme sample imbalance or long-tail problem, as seen in FSOD~\cite{Rethmeier20ltailFSL,Liu19LSLT}. Therefore, we aim to address this issue. FSCE~\cite{Sun21FSCE} and FSRC~\cite{Shangguan2022FSRC} have pointed out that the poor performance of FSOD is more related to inaccurate classification than inaccurate positioning. We have observed that this phenomenon occurs not only in Faster RCNN but also in DETR structures. Even with DETR, the focus remains on solving the problem of inaccurate classification. We argue that the extreme sample imbalance of FSOD results in the dominance of old knowledge from data-abundant classes in parameter optimization, even during fine-tuning. This means that the model will always have a certain bias toward the data-abundant classes. In order to overcome this problem, we have proposed a decoupled prompt module (DePrompt), which aims to add a prompt intervene at the low-dimensional feature stage of the model to enhance the model's concentration.
 This prompt module is decoupled from the old and novel classes, meaning that the two kinds of classes have independent prompt modules. Therefore, during training, the basic features learned by the model for old and new categories will not be mixed together, reducing the bias towards old categories.

In DETR, the encoder and decoder are responsible for feature encoding and decoding, respectively. This involves a process from shallow to deep and then back to shallow. In the traditional transformer structure, the feature transmission from the encoder to the decoder is linear, meaning that the decoder will only use the output of the last layer of the encoder as input. We hypothesize that this connection is inefficient because a shallow encoder might match a shallow decoder better, and vice versa. Therefore, we propose a method of skip connection between the encoder and decoder, which can effectively utilize the intermediate output of previous encoders at each decoder layer.
Furthermore, SQR~\cite{Chen23SQR} has pointed out that it is not only the final layer of the decoder that produces the correct prediction results, the output of the middle layer of the decoder sometimes produces better results. Therefore, we also conducted corresponding experiments in the FSOD scenario and indeed found that the intermediate layer of DETR predicted better results than the last layer when fine-tuning, as shown in ~\cref{fig:decoder_layers}. Therefore, we attempted to use the decoder output in an adaptive way to decide which layer to emphasize as the output. Specifically, we design an adaptive decoder fusion strategy so that the final output of the decoder module is determined by the weighting of the middle layer, instead of only relying on the output of the last layer. With the help of these modules, we could achieve significant improvement on the commonly used PASCAL VOC and COCO dataset, as shown in ~\cref{fig:intro}. Overall, our main contributions include:
\begin{itemize}
    \item We propose a decoupled prompt for novel and base categories that could effectively reduce the influence of the existing categories on the new class. This approach leads to the most significant and robust improvement.
    \item We discuss and simplify the skip connection between the encoder and decoder that does not require an extra learnable module and is competitive with the full skip connection.
	\item We design an unified adaptive decoder fusion strategy that dynamically determine the final output based on the weighting of all the middle decoder layers without the need for manual decoder layer selection tactics.
	\item Our approach has been demonstrated to be effective and reliable in both fine-tune and meta-learning paradigms, and our results achieve SOTA on meta-learning.
\end{itemize}

\section{Related works}
\label{sec:related works}

\begin{figure*}[ht]
  \centering
   \includegraphics[width=0.9\linewidth]{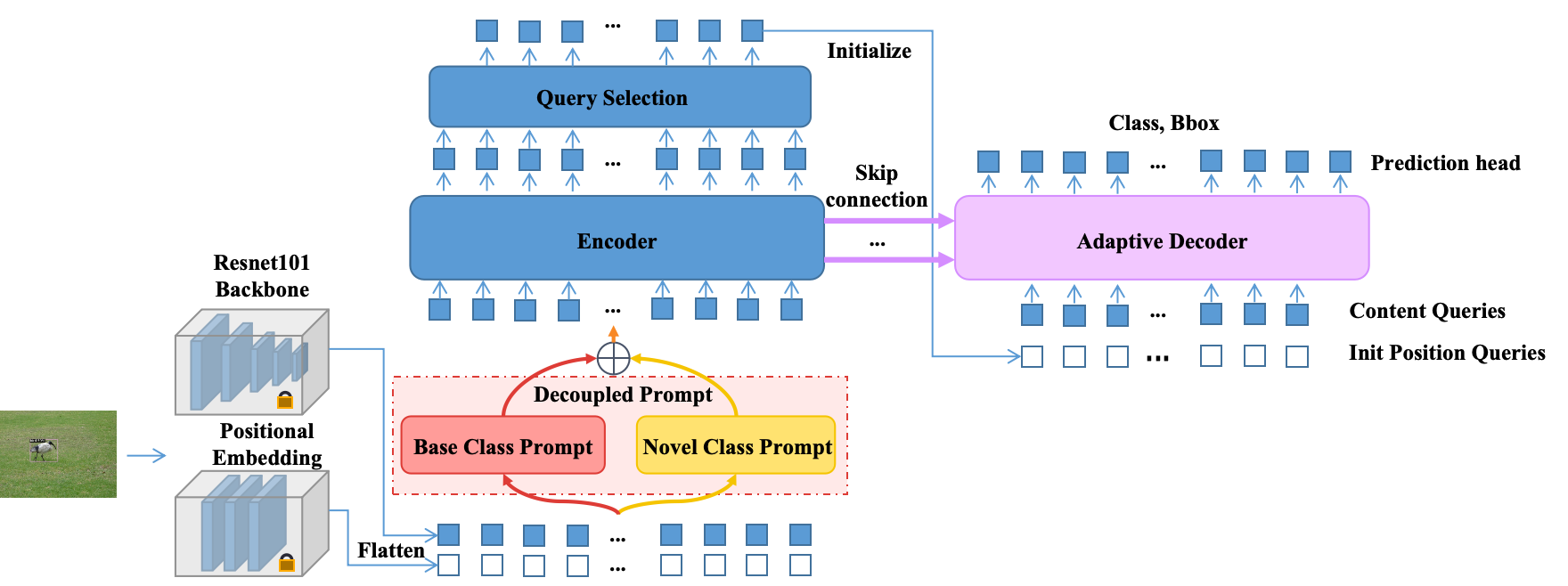}
   \caption{Overview of our DeDETR. Based on the DETR baseline, the decoupled prompt is marked in red block. The skip connection operator and the adaptive decoder are marked in pink.}
   \label{fig:DeDETR}
\end{figure*}

\subsection{Few-shot Object Detection}
\label{sec:Few-shot Object Detection}

Few-shot object detection (FSOD) has traditionally been divided into two paradigms: meta-learning and fine-tuning. In recent years, new paradigms have emerged that do not require fine-tuning~\cite{Bulat23FSDETR} and those that require retraining~\cite{Kaul22LVC}, both of which have shown positive results. In this paper, we conduct experiments using both the fine-tuning and meta-learning paradigms.
In the era of R-CNN~\cite{Girshick2013RichFH}, FSOD primarily relies on Faster R-CNN (FRCN)~\cite{Ren15FRCN} as the backbone. In addition to its success in standard FSOD settings, such as ~\cite{Chen2018LSTDAL,Karlinsky19RepMet,Wang19metaod,Fu19metassd,Kang19fsodrewei,Yang20Restor,Zhang21Hallu}, it and has made significant progress in other problem settings such as long-tail~\cite{Liu19LSLT,Li20imbalance}, cross-domain~\cite{Gao22cdod,Xiong23CDFSOD}, multi-modal~\cite{han2023multimodal,Yuan21TMDFS,Khoshboresh23mmfstd}. Later on, the emergence of the DETR~\cite{Carion20DETR} series has further expanded the options for FSOD~\cite{Zhang23MetaDETR,Bulat23FSDETR}.

TFA~\cite{Wang20TFA} has standardized the evaluation system of FSOD: during the fine-tuning stage, balanced data samples containing both old and new classes are used. Additionally, three different data divisions are implemented in VOC to evaluate the model's stability~\cite{Wang20TFA}. Meta-DETR~\cite{Zhang23MetaDETR} also adopts this category partitioning approach. However, unlike TFA which fine-tunes on the balanced samples, it fine-tunes the novel class while still strictly adhering to the $n$-shot settings, with the old class having more samples. In this paper, we also follow the evaluation system of Meta-DETR, using uneven fine-tuning data.
FSCE~\cite{Sun21FSCE} argues that classification is a more critical bottleneck than positioning. Our experiments on the DETR baseline confirm this, and therefore, we are also focusing more on addressing misclassification.
FSED~\cite{Guo23FSED} introduces a transformer-based class encoding approach to increase the inter-class distance, enabling the model to concentrate on the essential feature information.
Xu \etal~\cite{Xu23Generating} develops a generalized model using variational autoencoder to produce a large amount of augmented data.
Liu \etal ~\cite{Liu19LSLT} proposes that the few-shot problem is an extreme data imbalance issue, which is one of the causes of misclassification. We concur with this viewpoint and have developed our base-novel categories decoupled prompt module.


\subsection{Skip connection}

Skip connection was first introduced by ResNet~\cite{He26resnet} in order to alleviate the gradient vanishing problem, enabling information to bypass one or more layers and flow directly to the output. Later on, U-Net~\cite{Ronneberger15unet} incorporated a simple skip connection into the transformer, bridging the encoder and decoder to improve feature extraction. DSNet~\cite{Dai2019DenseSN} established a dense skip connection between the decoder and encoder to preserve information from previous scales. Lai \etal ~\cite{Lai2022RethinkingSC} enhanced the dense skip connection by incorporating spatial dimension features. While these dense skip connection performs well in most cases, they requires additional trainable weights. We reevaluate the original purpose of skip connections and propose that a relatively sparse skip connection, which does not involve all encoder layers in feature propagation, should be adequate.

\begin{figure*}[ht]
  \begin{subfigure}{0.27\linewidth}
    \centering
    \includegraphics[width=32mm]{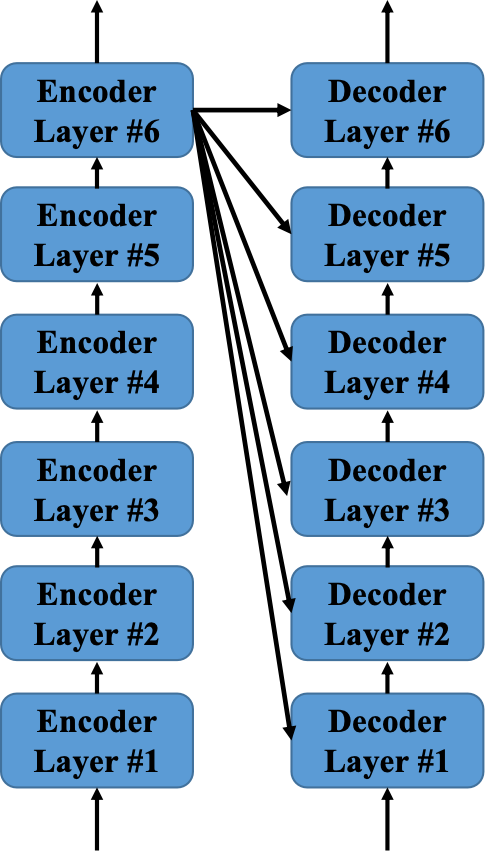}
    \caption{DETR baseline.}
    \label{fig:Origional DETR}
  \end{subfigure}
  \hfill
  \begin{subfigure}{0.27\linewidth}
    \centering
    \includegraphics[width=55mm]{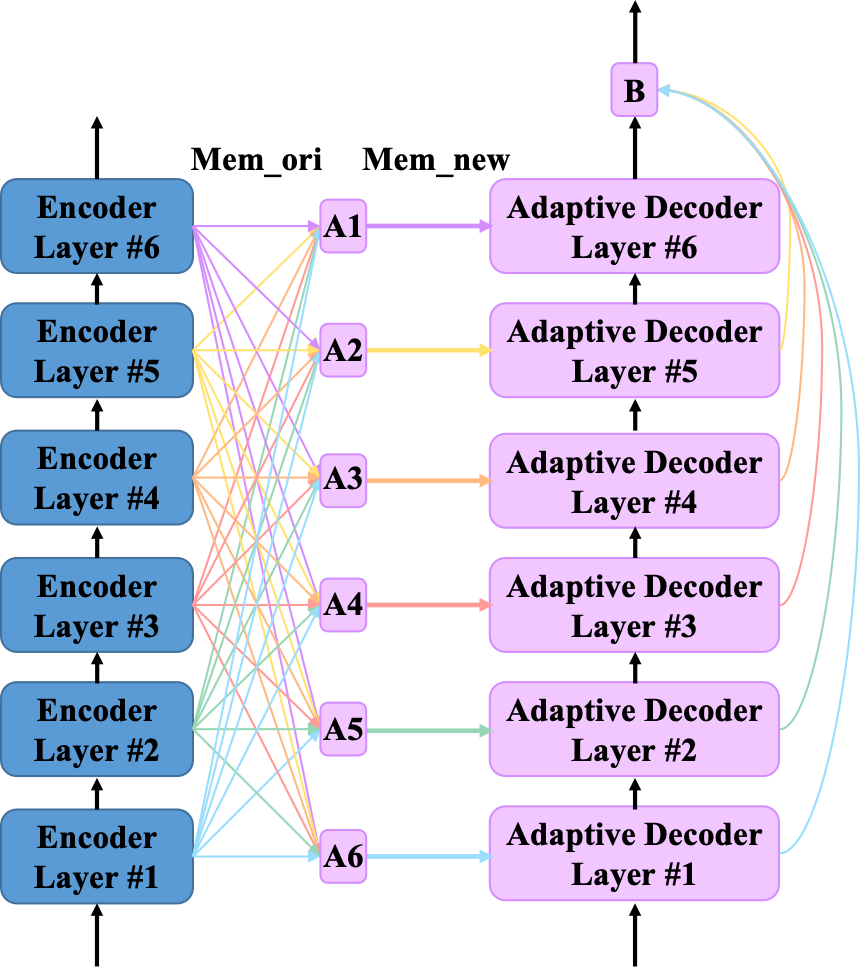}
    \caption{Learnable skip connection.}
    \label{fig:Learnable skip connection}
  \end{subfigure}
  \hfill
  \begin{subfigure}{0.27\linewidth}
    \centering
    \includegraphics[width=45mm]{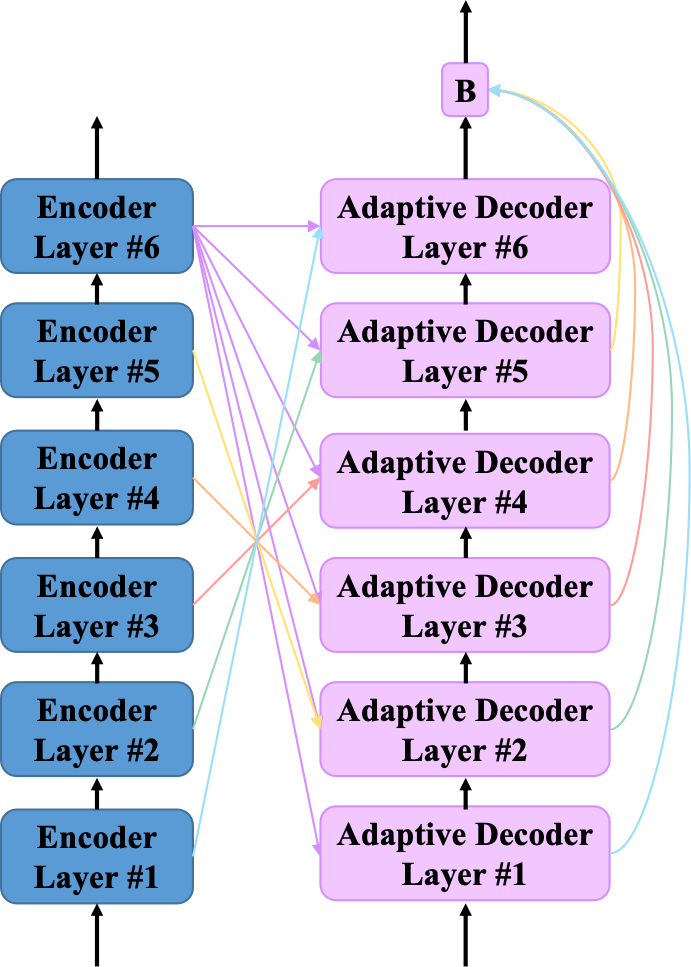}
    \caption{Soft skip connection.}
    \label{fig:Soft skip connection}
  \end{subfigure}
  \caption{Three type of connection between encoder and decoder. Block A1 to A6 indicates the parameters as in ~\cref{eq:learnable skip connection}. Block B refers to the parameters as in ~\cref{eq:adaptive decoder}.}
  \label{fig:Skip connection}
\end{figure*}

\subsection{Detection Transformer}

As a representation of generalized vision transformer, the detection transformer (DETR) first convert the traditional object detection task from an anchor-based method into an ensemble prediction problem that no longer needs hand-designed modules such as non-maximum suppression (NMS) and region proposals (RPN)~\cite{Carion20DETR}. Such integrated approach makes the end-to-end object detection task more intuitively perceptual. Additionally, its alignment with the transformer structure brings DETR more potential for expansion into visual-language multi-modal research~\cite{han2023multimodal}. Currently, the DETR based methods are becoming increasingly competitive against the R-CNN based methods. The variation works of DETR further promote the development of object detection and demonstrate competitive and even stronger generalizing ability comparing to R-CNN based methods~\cite{zhang2023dino}. Deformable DETR~\cite{zhu2021deformable} proposes a deformable attention module that significantly improves the perceptual ability of the model by focusing only on the sampling points near a reference point instead of all the sampling points. DAB-DETR~\cite{liu2022dabdetr} redesigns a 4D anchor to replace the 2D anchor points as the position queries. DN-DETR~\cite{Li2022dndetr} addresses the slow converging problem of DETR by adding a denoising loss that is trained with perturbed ground-truth labels. DINO~\cite{zhang2023dino} further improves the DN-DETR by applying contrastive learning and mixed query selection. Meta-DETR~\cite{Zhang23MetaDETR} applies DETR in the FSOD task for the first time under the meta-learning paradigm. FsDETR firstly tried to realize the non-retraining paradigm based on DETR~\cite{Bulat23FSDETR}. FsDETR~\cite{Bulat23FSDETR} explored few-shot DETR in a retraining-free manner. Since DINO has reached the best performance on general object detection task, we build our fine-tuning few-shot DeDETR baseline upon the DINO model. In addition, we build the meta-learning DeDETR model based on Meta-DETR. And we are the first ones to apply DETR under the fine-tuning paradigm.
As SQR~\cite{Chen23SQR} argues that the intermediate layers of the DETR decoder may yield better detection results compared to the last layer. Consequently, it establishes a connection between the current decoder layer and previous layers, allowing the decoder module to recollect previous decoder information. Additionally, DINO~\cite{zhang2023dino} designs a look forward twice structure to enable the current DETR decoder layer to connect with the subsequent two layers. These two strategies are similar but do not create a fully connected network among all decoder layers, and they do not offer a unified strategy for determining which layer should serve as the output. Therefore, we aim to design a unified decoder module that can automatically determine how to utilize information from different decoder layers.

\section{Proposed Methods}
\label{sec:proposed methods}

\subsection{Preliminary}
The task of few-shot object detection aims to first pre-train the model on the base classes ($C_B$) where there are sufficient training samples, and then fine-tuning it on both base and novel classes ($C_N$) that has only a few training samples for each class. Finally, the fine-tuned model is evaluated on the entire test dataset that includes $C_B \cup C_N$. In the PASCAL VOC few-shot training set, categories contains 1, 2, 3, 5, and 10 instances are named as 1 to 10-shots; 15 categories are selected as base classes, and the other 5 categories are considered novel classes. Similarly, in COCO datasets, the base classes contain 60 categories and the novel classes contain 20 categories under 10 and 30-shots setting. In the meta-learning paradigm, the $n$-shot support set means there are $n$ labeled instances from each few-shot category for training.

\subsection{Overview}
The overview of our method is in~\cref{fig:DeDETR}. The input image is first sent to both the feature pyramid network backbone and a positional embedding layer simultaneously to extract the visual and positional feature embedding. These flatten features are passed through our \textbf{decoupled prompts} to generate base-novel-categories-specified features according to the composition of the current training batch. Next, the decoupled prompt features are sent to the vanilla DETR encoder layers to obtain layer-wise memories and position queries. The encoder memories are then processed by our \textbf{skip connection operator} and serve as the input to the decoder layers. Finally, our proposed \textbf{adaptive decoder module} integrates each separate encoder layers to generate the final output features, which are sent to the prediction head consisting of two multilayer perceptrons (MLP) to obtain the classification and box regression results.

\subsection{Decoupled prompts (DePrompt)}

We suggest that during the baseline fine-tuning, the undiscriminating feature fusion of base class and novel class samples will decrease the affect from the novel class in terms of weights updating. Only a few samples from the novel class could hardly push a large model like DETR toward a suitable optimum without any specific operation. Therefore, we propose assigning separate weight sets to function as customized prompts for the novel and base classes, which we call decoupled prompts.

Specifically, we build two seperate deformable self-attention modules (structure from ~\cite{zhu2021deformable}, see Appendix) for the base and novel classes. These are added as input to the transformer encoder. The visual embedding and position embedding are sent to DePrompt and processed simultaneously through the base and novel prompt branches. We then check the sample composition of the current training batch and perform a conditional weighting operation: (Case 1) if the current training batch contains only base classes, the output features will come solely from the base prompt branch; (Case 2) if the current training batch contains only novel classes, the output features will come solely from the novel prompt branch; (Case 3) if the current batch contains both base and novel classes, the output features will be the weighted summation of base and novel prompt embedding; as shown in \cref{eq:DePrompt}, where $x$ is the input feature; $f_{DePrompt}(x)$ is the output feature; $f_{b_{pmt}}(x)$ and $f_{n_{pmt}}(x)$ are the base and novel prompt embedding respectively; $w$ is the weight of summation.

\begin{equation}
  f_{DePrompt}(x) = w * f_{b_{pmt}}(x) + (1-w) * f_{n_{pmt}}(x)
  \label{eq:DePrompt}
\end{equation}

\begin{table}[t]
\begin{center}
	\centering
	\resizebox{0.47\textwidth}{!}{
	\begin{tabular}{l|l|l|l|l}
		\toprule
		$w$ & \makecell{Training\\(Case1)} & \makecell{Training\\(Case2)} & \makecell{Training\\(Case3)} & Evaluating \\
		\midrule
		Hard & 1 & 0 & Fixed 0 $\sim$ 1 & Fixed 0 $\sim$ 1 \\
		Soft & 1 & 0 & $\frac{N_b}{N_b+N_n}$ & Fixed 0 $\sim$ 1 \\
		Learnable & 1 & 0 & Learnable & Learnable \\
		\bottomrule
	\end{tabular}
     }
\end{center}
\caption{The value selection strategies for $w$. $N_b$ and $N_n$ indicate the number of base and novel class instances respectively.}
\label{tab:DePrompt}
\end{table}

The value of $w$, as shown in \cref{tab:DePrompt}, is set to 1 and 0 for Case 1 and Case 2, respectively. For Case 3, we investigate three methods. (1) Hard coefficient: the weighting $w$ is fixed as a constant between 0 to 1 for both training and evaluating. (2) Soft coefficient: the weight $w$ is determined by the ratio of number of base instances and novel instances during training, and fixed as a constant during evaluating (here we set as 0.5 empirically). (3) Learnable coefficient: $w$ is a learnable parameter between 0 to 1 for both training and evaluating. Our experimental results indicate that the soft coefficient strategy performs best, please refer to ~\cref{sec:Effect of hyper-parameters}.

\begin{table*}[t]
\begin{center}
	\centering
	\resizebox{\textwidth}{!}{
		\begin{tabular}{l|l|lllll|lllll|lllll}
			\toprule
			\multirow{2}{*}{\diagbox[height=20pt,innerrightsep=25pt]{Method}{Shot}} & \multirow{2}{*}{Backbone} & \multicolumn{5}{c|}{Split1} & \multicolumn{5}{c|}{Split2} & \multicolumn{5}{c}{Split3} \\
			& & 1 & 2 & 3 & 5 & 10 & 1 & 2 & 3 & 5 & 10 & 1 & 2 & 3 & 5 & 10\\
			\midrule
			TFA w/ cos \cite{Wang20TFA} & FRCN-R101 & 39.8 & 36.1 & 44.7 & 55.7 & 56.0 & 23.5 & 26.9 & 34.1 & 35.1 & 39.1 & 30.8 & 34.8 & 42.8 & 49.5 & 49.8 \\
			MPSR \cite{Wu20MPSR} & FRCN-R101 & 41.7 & - & 51.4 & 55.2 & 61.8 & 24.4 & - & 39.2 & 39.9 & 47.8 & 35.6 & - & 42.3 & 48.0 & 49.7 \\
			FSCE \cite{Sun21FSCE} & FRCN-R101 & 44.2 & 43.8 & 51.4 & 61.9 & 63.4 & 27.3 & 29.5 & 43.5 & 44.2 & 50.2 & 37.2 & 41.9 & 47.5 & 54.6 & 58.5 \\
			Retentive R-CNN \cite{Fan21RetentiveRCNN} & FRCN-R101 & 42.4 & 45.8 & 45.9 & 53.7 & 56.1 & 21.7 & 27.8 & 35.2 & 37.0 & 40.3 & 30.2 & 37.6 & 43.0 & 49.7 & 50.1 \\
			DeFRCN \cite{Qiao21DeFRCN} & FRCN-R101 & 40.2 & 53.6 & 58.2 & 63.6 & 66.5 & 29.5 & \textcolor{blue}{\underline{39.7}} & 43.4 & 48.1 & 52.8 & 35.0 & 38.3 & 52.9 & 57.7 & \textcolor{blue}{\underline{60.8}} \\
			FSOD-UP \cite{Wu21FSODUP} & FRCN-R101 & 43.8 & 47.8 & 50.3 & 55.4 & 61.7 & 31.2 & 30.5 & 41.2 & 42.2 & 48.3 & 35.5 & 39.7 & 43.9 & 50.6 & 53.5 \\
			KFSOD \cite{Zhang22KFSOD} & FRCN-R101 & 44.6 & - & 54.4 & 60.9 & 65.8 & 37.8 & - & 43.1 & 48.1 & 50.4 & 34.8 & - & 44.1 & 52.7 & 53.9 \\
           FSRC \cite{Shangguan2022FSRC} & FRCN-R101 & 45.5 & 43.4 & 51.1 & 61.4 & 64.0 & 28.4 & 31.3 & 45.0 & 46.1 & 51.6 & 38.8 & 45.1 & 48.4 & 55.5 & 59.0 \\
           Pseudo-Labelling \cite{Kaul22LVC} & FRCN-R101 & \textbf{54.5} & 53.2 & 58.8 & 63.2 & 65.7 & 32.8 & 29.2 & \textbf{50.7} & 49.8 & 50.6 & 48.4 & \textcolor{blue}{\underline{52.7}} & \textcolor{blue}{\underline{55.0}} & 59.6 & 59.6\\
           Meta Faster R-CNN \cite{Han21MetaFRCN} & FRCN-R101 & 43.0 & 54.5 & 60.6 & \textbf{66.1} & 65.4 & 27.7 & 35.5 & 46.1 & 47.8 & 51.4 & 40.6 & 46.4 & 53.4 & \textcolor{blue}{\underline{59.9}} & 58.6 \\
			\midrule
			$\star$ DETR baseline (Our Impl.) & DETR-R101 & 19.4 & 29.4 & 35.0 & 49.8 & 54.7 & 11.1 & 20.8 & 20.4 & 30.4 & 40.6 & 10.6 & 23.9 & 31.9 & 37.6 & 46.3 \\
			\rowcolor{mygray} $\star$ \textbf{DeDETR (Our)} & DETR-R101 & 25.3 & 37.2 & 46.4 & 59.1 & 60.8 & 15.1 & 25.6 & 26.5 & 36.9 & 51.4 & 15.1 & 30.4 & 37.3 & 44.1 & 52.6 \\
			\midrule
			\midrule
	        $\dagger$ TIP \cite{Li21TIP} & FRCN-R101 & 27.7 & 36.5 & 43.3 & 50.2 & 59.6 & 22.7 & 30.1 & 33.8 & 40.9 & 46.9 & 21.7 & 30.6 & 38.1 & 44.5 & 50.9 \\
			$\dagger$ CME \cite{Li21CME} & FRCN-R101 & 41.5 & 47.5 & 50.4 & 58.2 & 60.9 & 27.2 & 30.2 & 41.4 & 42.5 & 46.8 & 34.3 & 39.6 & 45.1 & 48.3 & 51.5 \\
			$\dagger$ DC-Net \cite{Hu21DCNet} & FRCN-R101 & 33.9 & 37.4 & 43.7 & 51.1 & 59.6 & 23.2 & 24.8 & 30.6 & 36.7 & 46.6 & 32.3 & 34.9 & 39.7 & 42.6 & 50.7 \\
			$\dagger$ DeFRCN \cite{Qiao21DeFRCN} & FRCN-R101 & \textcolor{blue}{\underline{53.6}} & \textbf{57.5} & \textbf{61.5} & 64.1 & 60.8 & 30.1 & 38.1 & 47.0 & \textcolor{blue}{\underline{53.3}} & 47.9 & \textcolor{blue}{\underline{48.4}} & 50.9 & 52.3 & 54.9 & 57.4 \\
           $\dagger$ CGDP \cite{Li21CGDP} & FRCN-R2101 & 40.7 & 45.1 & 46.5 & 57.4 & 62.4 & 27.3 & 31.4 & 40.8 & 42.7 & 46.3 & 31.2 & 36.4 & 43.7 & 50.1 & 55.6 \\
           $\dagger$ Meta Faster R-CNN \cite{Han21MetaFRCN} & FRCN-R101 & 40.2 & 30.5 & 33.3 & 42.3 & 46.9 & 26.8 & 32.0 & 39.0 & 37.7 & 37.4 & 34.0 & 32.5 & 34.4 & 42.7 & 44.3 \\
           $\dagger$ FCT \cite{Han22FCT} & PVTv2-B2-Li & 49.9 & \textcolor{blue}{\underline{57.1}} & 57.9 & 63.2 & \textcolor{blue}{\underline{67.1}} & 27.6 & 34.5 & 43.7 & 49.2 & 51.2 & 39.5 & \textbf{54.7} & 52.3 & 57.0 & 58.7 \\
           $\star \dagger$ Meta-DETR \cite{Zhang23MetaDETR} & DETR-R101 & 40.6 & 51.4 & 58.0 & 59.2 & 63.6 & \textcolor{blue}{\underline{37.0}} & 36.6 & 43.7 & 49.1 & \textcolor{blue}{\underline{54.6}} & 41.6 & 45.9 & 52.7 & 58.9 & 60.6 \\
			\midrule
			\rowcolor{mygray} $\star \dagger$ \textbf{Meta-DeDETR (Our)} & DETR-R101 & 44.9 & 54.5 & \textcolor{blue}{\underline{61.1}} & \textcolor{blue}{\underline{65.1}} & \textbf{68.5} & \textbf{40.1} & \textbf{41.0} & \textcolor{blue}{\underline{47.5}} & \textbf{53.4} & \textbf{56.2} & \textbf{48.8} & 51.2 & \textbf{55.6} & \textbf{61.1} & \textbf{63.5} \\
			\bottomrule
		\end{tabular}
  }
\end{center}
\caption{Few-shot object detection performance on PASCAL VOC dataset. The novel classes nAP50 are evaluated on three separate splits. Our proposed DeDETR reaches new SOTA in most of the scenarios of meta-learning. \textbf{The highest nAP50 for each column} are in black bold text, and \textcolor{blue}{\underline{the second highest scores}} is in blue text with underline. Sign $\dagger$ indicates the meta-learning paradigm. Sign $\star$ indicates the utilization of imbalanced few-shot data set.}
\label{tab:voc}
\end{table*}

\subsection{Skip connection between encoder and decoder}

The encoder and decoder module of classic DETR are usually composed of 6 self-attention layers respectively. The output of the last layer of the encoder (\ie memory embedding) is the input of each decoder layers, as shown in ~\cref{fig:Origional DETR}. There have been a lot of works that discuss the possible ways of deeply connecting the encoder layers and decoder layers. As the transformer encoder encoding the low layer features to high layer features, and the decoder deciphers the high layer features back to low layer features, therefore, a skip connection between the encoder and decoder would be intuitive and trivial. Lai \etal proposes a skip connection between encode and decoder by collecting the outputs of all encoder layers and concatenating with the output of decoder layer in a weighted manner\cite{Lai2022RethinkingSC}. We follow this setting and explore a similar structure.

We explore two kinds of skip connection: learnable connection and soft connection. The learnable connection method contains a set of learnable parameters for the encoder output, as shown in ~\cref{eq:learnable skip connection}. For each decoder layer, the new input memory embedding will be the weighted combination of the original memory embedding from all encoder layers, as shown in ~\cref{fig:Learnable skip connection}, where $Mem\_new^{\{j\}}$ is the new encoder memory for decoder layer $j$, and $Mem\_ori^{\{i\}}$ is the original encoder memory from encoder layer $i$; $A_{ij}$ represents the normalized learnable parameter in 6 x 6 shape, each decoder layer $j$ has 6 parameters that weighting the $Out\_enc^{\{i\}}$

\begin{equation}
	Mem\_new^{\{j\}} = \sum_{i}^{i=6}A_{ij} * Mem\_ori^{\{i\}}
  \label{eq:learnable skip connection}
\end{equation}

For the soft skip connection, the new memory embedding only comes from one of the intermediate layers and the last layer of encoder, as shown in~\cref{fig:Skip connection}. For example, for an decoder layer $D_j$, the new memory embedding is a weighted summation of the last encoder layer $E_6$ and the corresponding intermediate layer $E_i$, where $i=6-j$. As shown in ~\cref{eq:soft skip connection}, in which $Mem\_new^{\{j\}}$ is the new encoder memory for decoder layer $j$; $Mem\_ori^{\{i\}}$ is the original encoder memory from encoder layer $i$; and $l_i$ represents the layer number (integer), from 0 to 5. Our experiments indicate that the soft skip connection has more advantages over the learnable skip connection, please refer to ~\cref{sec:Effect of hyper-parameters}.

{
\small
\begin{equation}
	Mem\_new^{\{j\}}\!=\!A\!*\!Mem\_ori^{\{6\}}+(1\!-\!A)\!*\!Mem\_ori^{\{i\}}
  \label{eq:soft skip connection}
\end{equation}
}

\subsection{Adaptive decoder selection}

As we mentioned in~\cref{sec:intro} and ~\cref{fig:decoder_layers}, the output of the 5 intermediate layers of decoder could possibly get better detection results than the last layer. Therefore, we intend to design a scheme that could let the model to determine which layer as the final output. Specifically, we design a set of learnable parameters that could be applied on weighting the decoder layers, as shown in ~\cref{fig:Soft skip connection,fig:Learnable skip connection}. In detail, we assign a set of normalized coefficients to integrate all of the decoder layer outputs, as shown in ~\cref{eq:adaptive decoder}, where $Dec\_new$ represents the new decoder output; $Dec\_ori^{\{j\}}$ is the original output from decoder layer $j$; and $B_{j}$ is the learnable coefficient for each decoder layer $j$.

\begin{equation}
	Dec\_new = \sum_{j}^{j=6}B_{j} * Dec\_ori^{\{j\}}
  \label{eq:adaptive decoder}
\end{equation}

\section{Experiments}

\subsection{Datasets}
Following previous works, we evaluate our few-shot object detection model on the two commonly used datasets: COCO and PASCAL VOC~\cite{Wang20TFA,Sun21FSCE,Qiao21DeFRCN,Zhang23MetaDETR}. For COCO dataset, 60 categories are selected as base categories for pre-training, while other 20 categories are novel categories. For PASCAL VOC dataset, 15 categories are base categories while the remaining 5 categories are defined as novel categories. Specifically, the few-shot PASCAL VOC dataset has three category splits for the purpose of eliminating the contingency while evaluating the model, each data split contains different base and novel categories combination.

\begin{table}[tb]
\centering
\scalebox{0.7}{
\begin{tabular}{l|l|ll|ll}
    \toprule
    \multirow{2}{*}{\diagbox[innerrightsep=30pt]{Method}{Shot}}& \multirow{2}{*}{Backbone}&\multicolumn{2}{c|}{Novel AP}&\multicolumn{2}{c}{Novel AP75}\\
    &&10&30&10&30\\
    \midrule
    TFA w/ cos \cite{Wang20TFA}&FRCN-R101&10.0&13.7&9.3&13.4\\
    FSCE \cite{Sun21FSCE}&FRCN-R101&11.9&16.4&10.5&16.2\\
    SVD \cite{WU2021Generalized}&FRCN-R101&12.0&16.0&10.4&15.3\\
    SRR-FSD \cite{Zhu21SRR-FSD}&FRCN-R101&11.3&14.7&9.8&13.5\\
    N-PME \cite{Liu22N-PME}&FRCN-R101&10.6&14.1&9.4&13.6\\
    FORD+BL \cite{Nguyen22FORDBL}&FRCN-R101&11.2&14.8&10.2&13.9\\
    FSRC \cite{Shangguan2022FSRC}&FRCN-R101&12.0&16.4&10.7&15.7\\
    Meta Faster R-CNN \cite{Han21MetaFRCN}&FRCN-R101 &12.7&16.6&10.8&15.8\\
    \midrule
    $\star$ DETR baseline (Our Impl.)&DETR-R101&6.3&10.2&5.9&9.1\\
    \rowcolor{mygray} $\star$ \textbf{DeDETR (Our)}&DETR-R101&10.6&14.3&10.2&14.1\\
    \midrule
    \midrule
    $\dagger$ FCT \cite{Han22FCT} & PVTv2-B2-Li&17.1&21.4&-&-\\
    $\dagger$ Meta Faster R-CNN \cite{Han21MetaFRCN}&FRCN-R101 &9.7&11.3&9.0&10.6\\
    $\star \dagger$ Meta-DETR \cite{Zhang23MetaDETR}&DETR-R101&\textcolor{blue}{\underline{19.0}}&\textcolor{blue}{\underline{22.2}}&\textcolor{blue}{\underline{19.7}}&\textcolor{blue}{\underline{22.8}}\\
    \midrule
    \rowcolor{mygray} $\star \dagger$ \textbf{Meta-DeDETR (Our)}&DETR-R101&\textbf{23.2}&\textbf{26.3}&\textbf{20.6}&\textbf{23.1}\\
    \bottomrule
\end{tabular}
}
\caption{Few-shot object detection performance on COCO dataset. Evaluation for novel classes AP and AP75 are listed. Our results have the highest scores above most previous works. \textbf{The highest AP for each column} is in black bold text, and \textcolor{blue}{\underline{the second highest scores}} are in regular text with underline. Sign $\dagger$ indicates the meta-learning paradigm. Sign $\star$ indicates the utilization of imbalanced few-shot data set.}
\label{tab:coco}
\end{table}

As we mentioned in ~\cref{sec:Few-shot Object Detection}, TFA~\cite{Wang20TFA} uses balanced $n$-shot base-novel data set where the number of instances for the novel and base classes are same; Meta-DETR~\cite{Zhang23MetaDETR} evaluates the model based on imbalanced data set where the number of instances for base classes is larger than $n$ (mostly less than $10 n$). In our experiment we follow the imbalanced fine-tuning data set from Meta-DETR.


\subsection{Training strategy}
We follow the same training strategy as Meta-DETR that uses ResNet-101 as the pre-trained backbone. Our DETR baseline is pre-trained on the base classes with no weights frozen. Then we fine-tune the model on few-shot novel and base classes, only freeze the ResNet-101 backbone. We run the training on 6 M40 GPUs with batch-size of 1 for fine-tuning and 4 for meta-learning paradigm. The position query is 900 for fine-tuning paradigm as in DINO~\cite{zhang2023dino}, and 300 for meta-learning paradigm as in Meta-DETR~\cite{Zhang23MetaDETR}. The training epoch is 60 with initial rate of 2e-4.

\begin{figure*}[t]
  \centering
   \includegraphics[width=0.9\linewidth]{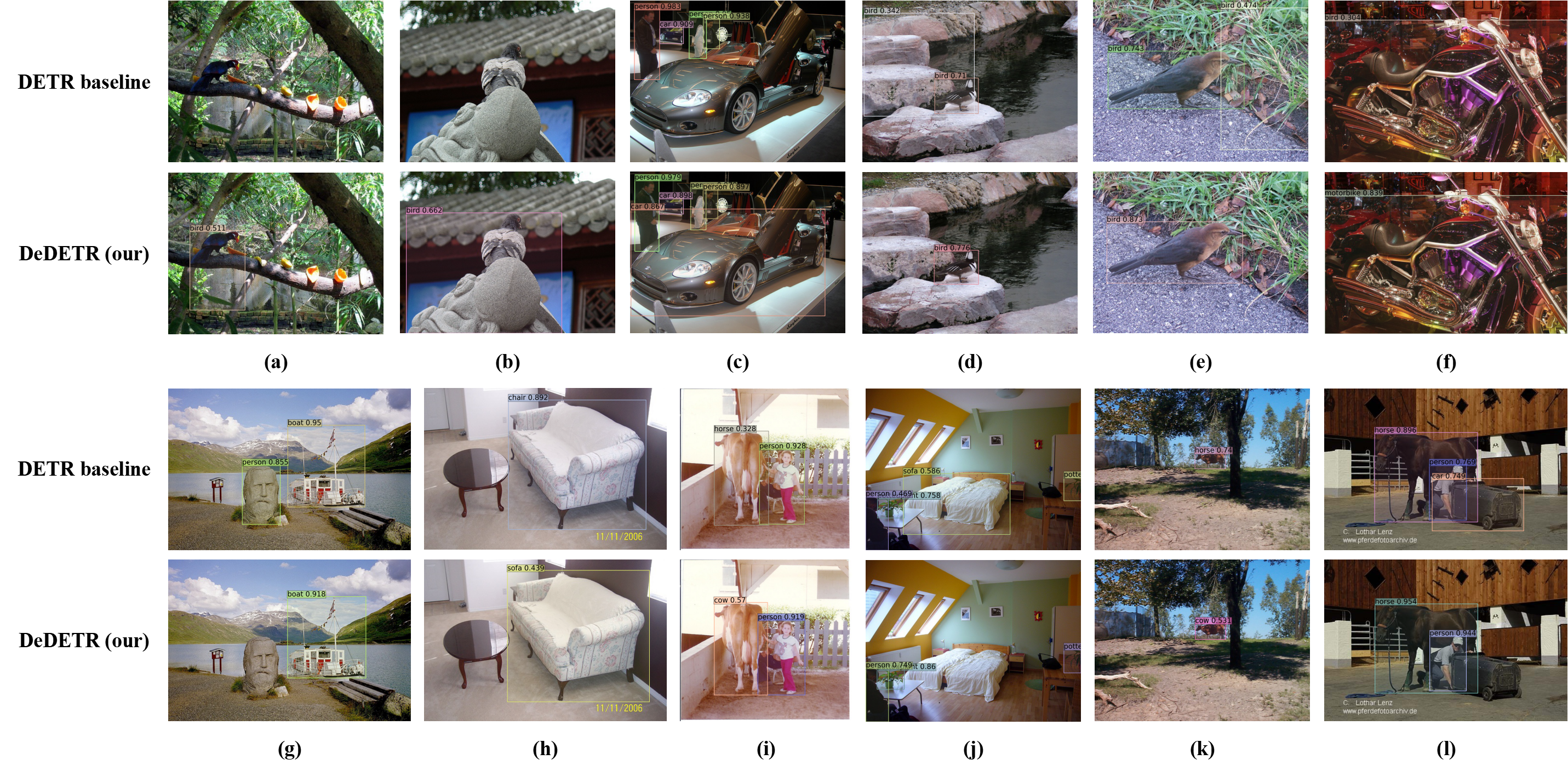}
   \caption{The visualization of our detection results comparing to the DETR baseline on PASCAL VOC test set.}
   \label{fig:detection results}
\end{figure*}

\subsection{Results on PASCAL VOC}
We present our experiment results on PASCAL VOC, as shown in~\cref{tab:voc}. We distinguish the methods based on fine-tuning and meta-learning. Also, we mark the evaluation scheme on balanced and imbalanced base-novel data sets.

For the meta-learning paradigm, we compared our method with previous SOTA, the results indicate that our method could outperform the previous works in most cases. For the fine-tuning paradigm, we not only report the results of our method, but also report our implementation of the DETR baseline on FSOD. Our results could outperform the baseline by up to 10\% in all cases.

Even though we could not beat the latest SOTA in the fine-tuning paradigm due to the less competitive DETR baseline we rely on, our result in the meta-learning paradigm could reach the SOTA. And more importantly, we could achieve significant improvement in both of these two paradigms, which could greatly demonstrate the generalization and robustness of our method.


\subsection{Results on COCO}
 
Our experimental results on COCO dataset are listed in ~\cref{tab:coco}. We evaluate our model on both fine-tuning and meta-learning paradigms, including AP and AP75 for the novel categories. We could observe that our method could get steady improvement on both fine-tuning and meta-learning networks, and we have reached the SOTA results.

\begin{table}[t]
	\centering
\scalebox{0.7}{
	\begin{tabular}{l|l|l}
		\toprule
		\multirow{2}{*}{Model} & \multicolumn{2}{c}{nAP50}\\
        & 1-shot & 5-shot \\
        \midrule
		DETR basl. & 19.4 & 49.8 \\
		DETR basl.+DePrompt  & 22.6 (+3.2) & 55.3 (+5.5) \\
		DETR basl.+DePrompt+Skip conn. & 24.1 (+1.5) & 57.5 (+2.2) \\
		DETR basl.+DePrompt+Skip conn.+Adpt. dec. & \textbf{25.2} (+1.1) & \textbf{59.1} (+1.6) \\
      \bottomrule
	\end{tabular}
 }
	\caption{Ablation study on our proposed modules based on the fine-tune DETR baseline (DETR basl.). The decoupled prompt (DePrompt) module gets the highest gain, up to 5\%; then the skip connection (Skip conn.) module and the adaptive decoder (Adpt. dec.) could get the second and third highest improvement respectively.}
	\label{tab:ablation study}
\end{table}

\subsection{Ablation study}

In this part, we mainly discuss the accuracy gain from each of our proposed three modules. The experiments are implemented on the PASCAL VOC 1-shot and 5-shot dataset based on the fine-tuning paradigm. As shown in \cref{tab:ablation study}, we accumulate our proposed modules on the DETR baseline. We want to highlight that our decoupled prompt provides the highest gain for the nAP50, while the skip connection module and the adaptive decoder module could achieve moderate improvement.

We assume that compared with the generalized class-agnostic feature extraction capability enhanced by skip connection and adaptive decoder, the decoupled prompt module can focus more on the distinction between novel and old categories. As mentioned in the introduction, the misclassification of FSOD to a large extent because it is easy to confuse some categories between the novel and old classes. However, our proposed decoupled prompt can effectively distinguish the feature embedding of the old and novel classes from the source by physically isolating them at the model weight level during training. Thus the maximum accuracy gain is achieved. This can be seen in more detailed experimental data. We take PASCAL VOC 5-shot split1 as an example and list the respective AP for each novel class, as shown in the ~\cref{tab:novel APs}, in which the improvement on 'bus' and 'motorbike' is prominent. This situation align with the analysis in FSRC~\cite{Shangguan2022FSRC} and FSCE~\cite{Sun21FSCE} that 'bus' and 'train' are easily confused, while 'motorbike' and 'bicycle' are easily confused. Such improvements can be also seen in ~\cref{fig:detection results}\textcolor{red}{f} to ~\cref{fig:detection results}\textcolor{red}{l}. Thus, our proposed decoupled prompt module is effective in improving the model's ability of recognizing novel classes.


\begin{table}[t]
	\centering
\scalebox{0.63}{
	\begin{tabular}{l|l|l|l|l|l}
		\toprule
	  Model  & Bird & Bus & Cow & Motorbike & Sofa\\ 
        \midrule
		DETR baseline & 42.4 & 57.7 & 66.1 & 48.6 & 34.3 \\
		DeDETR (our)  & 48.9 (+6.5) & \textbf{69.4 (+11.7)} & 74.1 (+8.0) & \textbf{61.8 (+13.2)} & 41.1 (+6.8)\\
         \bottomrule
	\end{tabular}
 }
	\caption{Class level comparison between our model and baseline.}
	\label{tab:novel APs}
\end{table}

\subsection{Effect of hyper-parameters}
\label{sec:Effect of hyper-parameters}

\textbf{Different coefficient $w$ for decoupled prompt} are list in ~\cref{tab:DePrompt_w}, the experiment is implemented base on PASCAL VOC 5-shot. We observe that the soft coefficient of $w$ could reach the highest nAP, while the hard and learnable coefficient of $w$ are weaker. The principal differences are: when a training batch contains both base and novel samples, the hard and learnable coefficients are unable perceive the ratio of novel and base samples directly, and therefore hard to assign the proper gradient to the novel and base prompt respectively, which makes the model harder to converge. However, the soft coefficient directly assign the loss energy according to the number of samples, which could achieve better convergence.

\textbf{Comparison between soft and learnable skip connection} are listed in ~\cref{tab:skip connection}. We implement this experiment on PASCAL VOC 1-shot and 5-shot, and we could observe that the gap between the soft and learnable skip connection is negligible. Therefore, to some extent, as the input of the $i^{th}$ decoder layer, the weighted combination of the last layer and corresponding $(6-i)^{th}$ layer of the encoder is sufficient. Even though the learnable full skip connection that utilize all of the encoder layer could achieve higher AP50, considering the newly introduced extra model parameters, such marginal improvement is not a desirable trade-off. Therefore, we recommend the future works to use our explored soft skip connection, which is simple yet effective.

\subsection{Detection results}

We run the inference on the PASCAL VOC test set, as seen in ~\cref{fig:detection results}. The confidential threshold is set as 0.3 for all images. ~\cref{fig:detection results}\textcolor{red}{a} to ~\cref{fig:detection results}\textcolor{red}{c} have shown our improvement on the missing detection. ~\cref{fig:detection results}\textcolor{red}{d} and ~\cref{fig:detection results}\textcolor{red}{e} demonstrate our improvement on the incorrect box regression. ~\cref{fig:detection results}\textcolor{red}{f} to ~\cref{fig:detection results}\textcolor{red}{l} indicate our improvement on the misclassification.


\begin{table}
\begin{center}
\scalebox{0.7}{
	\centering
	\begin{tabular}{l|llllll}
		\toprule
		$w$ & 0.0 & 0.2 & 0.4 & 0.6 & 0.8 & 1.0\\
		\midrule
		Hard (train and eval) & 39.7 & 53.4 & 52.9 & 39.6 & 38.7 & 35.2\\
		Soft (eval) & 35.4 & 41.3 & 47.2 & \textbf{55.9} & 51.5 & 48.8\\
		Learnable & \multicolumn{6}{c}{40.2} \\
		\bottomrule
	\end{tabular}
 }
\end{center}
\caption{Effect of different $w$ strategies.}
\label{tab:DePrompt_w}
\end{table}

\begin{table}
\begin{center}
\scalebox{0.8}{
	\centering
	\begin{tabular}{l|l|l}
		\toprule
	     & Soft & Learnable \\
		\midrule
		1-shot nAP & 24.0 & 24.1 (+0.1)\\
		5-shot nAP & 57.3 & 57.5 (+0.2)\\
		\bottomrule
	\end{tabular}
 }
\end{center}
\caption{Comparison to the soft and learnable skip connection.}
\label{tab:skip connection}
\end{table}

\section{Conclusion}
To further improve the accuracy of few-shot object detection, we proposes improvements focusing on sample imbalance and feature propagation.
Our decoupled prompt module demonstrates that the weight separation strategy effectively alleviates bias from data-rich classes.
Additionally, we introduce a simplified soft skip connection between the encoder and decoder, which competes effectively with the dense skip connection.
Furthermore, we propose to effectively utilize each decoder layer by fusing the intermediate decoder layers adaptively as the output.
Tests conducted on widely-used datasets such as PASCAL VOC and MSCOCO have consistently shown a 5-10\% performance increase in our model, making it superior to contemporary models.

{
    \small
    \bibliographystyle{ieeenat_fullname}
    \bibliography{main}

\begin{thebibliography}{51}
\providecommand{\natexlab}[1]{#1}
\providecommand{\url}[1]{\texttt{#1}}
\expandafter\ifx\csname urlstyle\endcsname\relax
  \providecommand{\doi}[1]{doi: #1}\else
  \providecommand{\doi}{doi: \begingroup \urlstyle{rm}\Url}\fi

\bibitem[Bochkovskiy et~al.(2020)Bochkovskiy, Wang, and
  Liao]{Bochkovskiy20YOLOv4}
Alexey Bochkovskiy, Chien-Yao Wang, and Hong-Yuan~Mark Liao.
\newblock Yolov4: Optimal speed and accuracy of object detection.
\newblock \emph{arXiv: Computer Vision and Pattern Recognition}, 2020.

\bibitem[Bulat et~al.(2023)Bulat, Guerrero, Martinez, and
  Tzimiropoulos]{Bulat23FSDETR}
Adrian Bulat, Ricardo Guerrero, Brais Martinez, and Georgios Tzimiropoulos.
\newblock Fs-detr: Few-shot detection transformer with prompting and without
  re-training.
\newblock In \emph{Proceedings of the IEEE/CVF International Conference on
  Computer Vision (ICCV)}, pages 11793--11802, 2023.

\bibitem[Carion et~al.(2020)Carion, Massa, Synnaeve, Usunier, Kirillov, and
  Zagoruyko]{Carion20DETR}
Nicolas Carion, Francisco Massa, Gabriel Synnaeve, Nicolas Usunier, Alexander
  Kirillov, and Sergey Zagoruyko.
\newblock End-to-end object detection with transformers.
\newblock In \emph{Computer Vision -- ECCV 2020}, pages 213--229, Cham, 2020.
  Springer International Publishing.

\bibitem[Chen et~al.(2023)Chen, Zhang, Hu, Huang, Zhu, and Savvides]{Chen23SQR}
Fangyi Chen, Han Zhang, Kai Hu, Yu-Kai Huang, Chenchen Zhu, and Marios
  Savvides.
\newblock Enhanced training of query-based object detection via selective query
  recollection.
\newblock In \emph{2023 IEEE/CVF Conference on Computer Vision and Pattern
  Recognition (CVPR)}, pages 23756--23765, 2023.

\bibitem[Chen et~al.(2018)Chen, Wang, Wang, and Qiao]{Chen2018LSTDAL}
Hao Chen, Yali Wang, Guoyou Wang, and Yu Qiao.
\newblock Lstd: A low-shot transfer detector for object detection.
\newblock In \emph{AAAI Conference on Artificial Intelligence}, 2018.

\bibitem[Dai et~al.(2019)Dai, Liu, Ma, Cao, Zhao, and Zhang]{Dai2019DenseSN}
Feng Dai, Hao Liu, Yike Ma, Juan Cao, Qiang Zhao, and Yongdong Zhang.
\newblock Dense scale network for crowd counting.
\newblock \emph{Proceedings of the 2021 International Conference on Multimedia
  Retrieval}, 2019.

\bibitem[Fan et~al.(2021)Fan, Ma, Li, and Sun]{Fan21RetentiveRCNN}
Zhibo Fan, Yuchen Ma, Zeming Li, and Jian Sun.
\newblock Generalized few-shot object detection without forgetting.
\newblock In \emph{CVPR}, pages 4525--4534, 2021.

\bibitem[Fu et~al.(2019)Fu, Zhang, Zhang, Yan, Chang, Zhang, and
  Sun]{Fu19metassd}
Kun Fu, Tengfei Zhang, Yue Zhang, Menglong Yan, Zhonghan Chang, Zhengyuan
  Zhang, and Xian Sun.
\newblock Meta-ssd: Towards fast adaptation for few-shot object detection with
  meta-learning.
\newblock \emph{{IEEE} Access}, 7:\penalty0 77597--77606, 2019.

\bibitem[Gao et~al.(2022)Gao, Yang, Huang, Xie, Li, and Zheng]{Gao22cdod}
Yipeng Gao, Lingxiao Yang, Yunmu Huang, Song Xie, Shiyong Li, and Wei-Shi
  Zheng.
\newblock Acrofod: An adaptive method for cross-domain few-shot object
  detection.
\newblock In \emph{Computer Vision -- ECCV 2022}, pages 673--690, Cham, 2022.
  Springer Nature Switzerland.

\bibitem[Girshick et~al.(2013)Girshick, Donahue, Darrell, and
  Malik]{Girshick2013RichFH}
Ross~B. Girshick, Jeff Donahue, Trevor Darrell, and Jitendra Malik.
\newblock Rich feature hierarchies for accurate object detection and semantic
  segmentation.
\newblock \emph{CoRR}, abs/1311.2524, 2013.

\bibitem[Guo et~al.(2023)Guo, Yang, Wei, Ye, and Zhang]{Guo23FSED}
Xueqiang Guo, Hanqing Yang, Mohan Wei, Xiaotong Ye, and Yu Zhang.
\newblock Few-shot object detection via class encoding and multi-target
  decoding.
\newblock \emph{IET Cyber-Systems and Robotics}, 5\penalty0 (2):\penalty0
  e12088, 2023.

\bibitem[Han et~al.(2021)Han, Huang, Ma, He, and Chang]{Han21MetaFRCN}
Guangxing Han, Shiyuan Huang, Jiawei Ma, Yicheng He, and Shih{-}Fu Chang.
\newblock Meta faster {R-CNN:} towards accurate few-shot object detection with
  attentive feature alignment.
\newblock \emph{CoRR}, abs/2104.07719, 2021.

\bibitem[Han et~al.(2022)Han, Ma, Huang, Chen, and Chang]{Han22FCT}
Guangxing Han, Jiawei Ma, Shiyuan Huang, Long Chen, and Shih-Fu Chang.
\newblock Few-shot object detection with fully cross-transformer.
\newblock In \emph{2022 IEEE/CVF Conference on Computer Vision and Pattern
  Recognition (CVPR)}, pages 5311--5320, 2022.

\bibitem[Han et~al.(2023)Han, Chen, Ma, Huang, Chellappa, and
  Chang]{han2023multimodal}
Guangxing Han, Long Chen, Jiawei Ma, Shiyuan Huang, Rama Chellappa, and Shih-Fu
  Chang.
\newblock Multi-modal few-shot object detection with meta-learning-based
  cross-modal prompting, 2023.

\bibitem[He et~al.(2016)He, Zhang, Ren, and Sun]{He26resnet}
Kaiming He, Xiangyu Zhang, Shaoqing Ren, and Jian Sun.
\newblock Deep residual learning for image recognition.
\newblock In \emph{2016 IEEE Conference on Computer Vision and Pattern
  Recognition (CVPR)}, pages 770--778, 2016.

\bibitem[Hu et~al.(2021)Hu, Bai, Li, Cui, and Wang]{Hu21DCNet}
Hanzhe Hu, Shuai Bai, Aoxue Li, Jinshi Cui, and Liwei Wang.
\newblock Dense relation distillation with context-aware aggregation for
  few-shot object detection.
\newblock In \emph{2021 IEEE/CVF Conference on Computer Vision and Pattern
  Recognition (CVPR)}, pages 10180--10189, 2021.

\bibitem[Kang et~al.(2019)Kang, Liu, Wang, Yu, Feng, and
  Darrell]{Kang19fsodrewei}
Bingyi Kang, Zhuang Liu, Xin Wang, Fisher Yu, Jiashi Feng, and Trevor Darrell.
\newblock Few-shot object detection via feature reweighting.
\newblock In \emph{2019 {IEEE/CVF} International Conference on Computer Vision,
  {ICCV} 2019, Seoul, Korea (South), October 27 - November 2, 2019}, pages
  8419--8428. {IEEE}, 2019.

\bibitem[Karlinsky et~al.(2019)Karlinsky, Shtok, Harary, Schwartz, Aides,
  Feris, Giryes, and Bronstein]{Karlinsky19RepMet}
Leonid Karlinsky, Joseph Shtok, Sivan Harary, Eli Schwartz, Amit Aides, Rogerio
  Feris, Raja Giryes, and Alex~M. Bronstein.
\newblock Repmet: Representative-based metric learning for classification and
  few-shot object detection.
\newblock In \emph{Proceedings of the IEEE/CVF Conference on Computer Vision
  and Pattern Recognition (CVPR)}, 2019.

\bibitem[Kaul et~al.(2022)Kaul, Xie, and Zisserman]{Kaul22LVC}
Prannay Kaul, Weidi Xie, and Andrew Zisserman.
\newblock Label, verify, correct: A simple few shot object detection method.
\newblock In \emph{2022 IEEE/CVF Conference on Computer Vision and Pattern
  Recognition (CVPR)}, pages 14217--14227, 2022.

\bibitem[Khoshboresh-Masouleh and Shah-Hosseini(2023)]{Khoshboresh23mmfstd}
Mehdi Khoshboresh-Masouleh and Reza Shah-Hosseini.
\newblock Multimodal few-shot target detection based on uncertainty analysis in
  time-series images.
\newblock \emph{Drones}, 7\penalty0 (2), 2023.

\bibitem[Lai et~al.(2022)Lai, Sun, Tian, Ding, Wu, and
  Wang]{Lai2022RethinkingSC}
Zhitong Lai, Haichao Sun, Rui Tian, Nannan Ding, Zhiguo Wu, and Yanjie Wang.
\newblock Rethinking skip connections in encoder-decoder networks for monocular
  depth estimation.
\newblock \emph{ArXiv}, abs/2208.13441, 2022.

\bibitem[Li and Li(2021)]{Li21TIP}
Aoxue Li and Zhenguo Li.
\newblock Transformation invariant few-shot object detection.
\newblock In \emph{2021 IEEE/CVF Conference on Computer Vision and Pattern
  Recognition (CVPR)}, pages 3093--3101, 2021.

\bibitem[Li et~al.(2021{\natexlab{a}})Li, Yang, Liu, Liu, Ji, and Ye]{Li21CME}
Bohao Li, Boyu Yang, Chang Liu, Feng Liu, Rongrong Ji, and Qixiang Ye.
\newblock Beyond max-margin: Class margin equilibrium for few-shot object
  detection.
\newblock In \emph{2021 IEEE/CVF Conference on Computer Vision and Pattern
  Recognition (CVPR)}, pages 7359--7368, 2021{\natexlab{a}}.

\bibitem[Li et~al.(2022)Li, Zhang, Liu, Guo, Ni, and Zhang]{Li2022dndetr}
Feng Li, Hao Zhang, Shilong Liu, Jian Guo, Lionel~M. Ni, and Lei Zhang.
\newblock Dn-detr: Accelerate detr training by introducing query denoising.
\newblock In \emph{2022 IEEE/CVF Conference on Computer Vision and Pattern
  Recognition (CVPR)}, pages 13609--13617, 2022.

\bibitem[Li et~al.(2020)Li, Wang, Kang, Tang, Wang, Li, and
  Feng]{Li20imbalance}
Yu Li, Tao Wang, Bingyi Kang, Sheng Tang, Chunfeng Wang, Jintao Li, and Jiashi
  Feng.
\newblock Overcoming classifier imbalance for long-tail object detection with
  balanced group softmax.
\newblock In \emph{2020 IEEE/CVF Conference on Computer Vision and Pattern
  Recognition (CVPR)}, pages 10988--10997, 2020.

\bibitem[Li et~al.(2021{\natexlab{b}})Li, Zhu, Cheng, Wang, Teo, Xiang,
  Vadakkepat, and Lee]{Li21CGDP}
Yiting Li, Haiyue Zhu, Yu Cheng, Wenxin Wang, Chek~Sing Teo, Cheng Xiang,
  Prahlad Vadakkepat, and Tong~Heng Lee.
\newblock Few-shot object detection via classification refinement and
  distractor retreatment.
\newblock In \emph{2021 IEEE/CVF Conference on Computer Vision and Pattern
  Recognition (CVPR)}, pages 15390--15398, 2021{\natexlab{b}}.

\bibitem[Liu et~al.(2022{\natexlab{a}})Liu, Li, Zhang, Yang, Qi, Su, Zhu, and
  Zhang]{liu2022dabdetr}
Shilong Liu, Feng Li, Hao Zhang, Xiao Yang, Xianbiao Qi, Hang Su, Jun Zhu, and
  Lei Zhang.
\newblock {DAB}-{DETR}: Dynamic anchor boxes are better queries for {DETR}.
\newblock In \emph{International Conference on Learning Representations},
  2022{\natexlab{a}}.

\bibitem[Liu et~al.(2022{\natexlab{b}})Liu, Wang, Yu, Tao, Wang, and
  Wu]{Liu22N-PME}
Weijie Liu, Chong Wang, Shenghao Yu, Chenchen Tao, Jun Wang, and Jiafei Wu.
\newblock Novel instance mining with pseudo-margin evaluation for few-shot
  object detection.
\newblock \emph{ICASSP 2022 - 2022 IEEE International Conference on Acoustics,
  Speech and Signal Processing (ICASSP)}, pages 2250--2254, 2022{\natexlab{b}}.

\bibitem[Liu et~al.(2019)Liu, Miao, Zhan, Wang, Gong, and Yu]{Liu19LSLT}
Ziwei Liu, Zhongqi Miao, Xiaohang Zhan, Jiayun Wang, Boqing Gong, and Stella~X.
  Yu.
\newblock Large-scale long-tailed recognition in an open world.
\newblock In \emph{2019 IEEE/CVF Conference on Computer Vision and Pattern
  Recognition (CVPR)}, pages 2532--2541, 2019.

\bibitem[Qiao et~al.(2021)Qiao, Zhao, Li, Qiu, Wu, and Zhang]{Qiao21DeFRCN}
Limeng Qiao, Yuxuan Zhao, Zhiyuan Li, Xi Qiu, Jianan Wu, and Chi Zhang.
\newblock Defrcn: Decoupled faster r-cnn for few-shot object detection.
\newblock In \emph{2021 IEEE/CVF International Conference on Computer Vision
  (ICCV)}, pages 8661--8670, 2021.

\bibitem[Ren et~al.(2015)Ren, He, Girshick, and Sun]{Ren15FRCN}
Shaoqing Ren, Kaiming He, Ross~B. Girshick, and Jian Sun.
\newblock Faster {R-CNN:} towards real-time object detection with region
  proposal networks.
\newblock In \emph{Advances in Neural Information Processing Systems 28: Annual
  Conference on Neural Information Processing Systems 2015, December 7-12,
  2015, Montreal, Quebec, Canada}, pages 91--99, 2015.

\bibitem[Rethmeier and Augenstein(2020)]{Rethmeier20ltailFSL}
Nils Rethmeier and Isabelle Augenstein.
\newblock Long-tail zero and few-shot learning via contrastive pretraining on
  and for small data.
\newblock \emph{CoRR}, abs/2010.01061, 2020.

\bibitem[Ronneberger et~al.(2015)Ronneberger, Fischer, and
  Brox]{Ronneberger15unet}
Olaf Ronneberger, Philipp Fischer, and Thomas Brox.
\newblock U-net: Convolutional networks for biomedical image segmentation.
\newblock In \emph{Medical Image Computing and Computer-Assisted Intervention
  -- MICCAI 2015}, pages 234--241, Cham, 2015. Springer International
  Publishing.

\bibitem[Shangguan et~al.(2022)Shangguan, Huai, Liu, and
  Jiang]{Shangguan2022FSRC}
Zeyu Shangguan, Lian Huai, Tong Liu, and Xingqun Jiang.
\newblock Few-shot object detection with refined contrastive learning.
\newblock \emph{ArXiv}, abs/2211.13495, 2022.

\bibitem[Sun et~al.(2021)Sun, Li, Cai, Yuan, and Zhang]{Sun21FSCE}
Bo Sun, Banghuai Li, Shengcai Cai, Ye Yuan, and Chi Zhang.
\newblock Fsce: Few-shot object detection via contrastive proposal encoding.
\newblock In \emph{Proceedings of the IEEE/CVF Conference on Computer Vision
  and Pattern Recognition (CVPR)}, pages 7352--7362, 2021.

\bibitem[Vu et~al.(2022)Vu, Nguyen, Nguyen, Nguyen, Ngo, Do, and
  Nguyen]{Nguyen22FORDBL}
Anh-Khoa~Nguyen Vu, Nhat-Duy Nguyen, Khanh-Duy Nguyen, Vinh-Tiep Nguyen,
  Thanh~Duc Ngo, Thanh-Toan Do, and Tam~V. Nguyen.
\newblock Few-shot object detection via baby learning.
\newblock \emph{Image and Vision Computing}, 120:\penalty0 104398, 2022.

\bibitem[Wang et~al.(2020)Wang, Huang, Darrell, Gonzalez, and Yu]{Wang20TFA}
Xin Wang, Thomas~E Huang, Trevor Darrell, Joseph~E Gonzalez, and Fisher Yu.
\newblock Frustratingly simple few-shot object detection.
\newblock \emph{ICML}, 2020.

\bibitem[Wang et~al.(2019)Wang, Ramanan, and Hebert]{Wang19metaod}
Yu-Xiong Wang, Deva Ramanan, and Martial Hebert.
\newblock Meta-learning to detect rare objects.
\newblock In \emph{2019 IEEE/CVF International Conference on Computer Vision
  (ICCV)}, pages 9924--9933, 2019.

\bibitem[Wu et~al.(2021)Wu, Han, Zhu, and Yang]{Wu21FSODUP}
Aming Wu, Yahong Han, Linchao Zhu, and Yi Yang.
\newblock Universal-prototype enhancing for few-shot object detection.
\newblock In \emph{2021 {IEEE/CVF} International Conference on Computer Vision,
  {ICCV} 2021, Montreal, QC, Canada, October 10-17, 2021}, pages 9547--9556.
  {IEEE}, 2021.

\bibitem[WU et~al.(2021)WU, Zhao, Deng, and Liu]{WU2021Generalized}
Aming WU, Suqi Zhao, Cheng Deng, and Wei Liu.
\newblock Generalized and discriminative few-shot object detection via
  svd-dictionary enhancement.
\newblock In \emph{Advances in Neural Information Processing Systems}, pages
  6353--6364. Curran Associates, Inc., 2021.

\bibitem[Wu et~al.(2020)Wu, Liu, Huang, and Wang]{Wu20MPSR}
Jiaxi Wu, Songtao Liu, Di Huang, and Yunhong Wang.
\newblock Multi-scale positive sample refinement for few-shot object detection.
\newblock In \emph{European Conference on Computer Vision}, 2020.

\bibitem[Xiong(2023)]{Xiong23CDFSOD}
Wuti Xiong.
\newblock Cd-fsod: A benchmark for cross-domain few-shot object detection.
\newblock In \emph{ICASSP 2023 - 2023 IEEE International Conference on
  Acoustics, Speech and Signal Processing (ICASSP)}, pages 1--5, 2023.

\bibitem[Xu et~al.(2023)Xu, Le, and Samaras]{Xu23Generating}
Jingyi Xu, Hieu Le, and Dimitris Samaras.
\newblock Generating features with increased crop-related diversity for
  few-shot object detection.
\newblock In \emph{2023 IEEE/CVF Conference on Computer Vision and Pattern
  Recognition (CVPR)}, pages 19713--19722, 2023.

\bibitem[Yang et~al.(2020)Yang, Wei, Shi, and Li]{Yang20Restor}
Yukuan Yang, Fangyun Wei, Miaojing Shi, and Guoqi Li.
\newblock Restoring negative information in few-shot object detection.
\newblock In \emph{Advances in Neural Information Processing Systems 33: Annual
  Conference on Neural Information Processing Systems 2020, NeurIPS 2020,
  December 6-12, 2020, virtual}, 2020.

\bibitem[Yuan et~al.(2021)Yuan, Duan, Wang, and En]{Yuan21TMDFS}
Ying Yuan, Lijuan Duan, Wenjian Wang, and Qing En.
\newblock Tmd-fs: Improving few-shot object detection with transformer
  multi-modal directing.
\newblock In \emph{Pattern Recognition and Computer Vision}, pages 447--458,
  Cham, 2021. Springer International Publishing.

\bibitem[Zhang et~al.(2023{\natexlab{a}})Zhang, Luo, Cui, Lu, and
  Xing]{Zhang23MetaDETR}
Gongjie Zhang, Zhipeng Luo, Kaiwen Cui, Shijian Lu, and Eric~P. Xing.
\newblock Meta-detr: Image-level few-shot detection with inter-class
  correlation exploitation.
\newblock \emph{IEEE Transactions on Pattern Analysis and Machine
  Intelligence}, 45\penalty0 (11):\penalty0 12832--12843, 2023{\natexlab{a}}.

\bibitem[Zhang et~al.(2023{\natexlab{b}})Zhang, Li, Liu, Zhang, Su, Zhu, Ni,
  and Shum]{zhang2023dino}
Hao Zhang, Feng Li, Shilong Liu, Lei Zhang, Hang Su, Jun Zhu, Lionel Ni, and
  Heung-Yeung Shum.
\newblock {DINO}: {DETR} with improved denoising anchor boxes for end-to-end
  object detection.
\newblock In \emph{The Eleventh International Conference on Learning
  Representations}, 2023{\natexlab{b}}.

\bibitem[Zhang et~al.(2022)Zhang, Wang, Murray, and Koniusz]{Zhang22KFSOD}
Shan Zhang, Lei Wang, Naila Murray, and Piotr Koniusz.
\newblock Kernelized few-shot object detection with efficient integral
  aggregation.
\newblock In \emph{2022 IEEE/CVF Conference on Computer Vision and Pattern
  Recognition (CVPR)}, pages 19185--19194, 2022.

\bibitem[Zhang and Wang(2021)]{Zhang21Hallu}
Weilin Zhang and Yu{-}Xiong Wang.
\newblock Hallucination improves few-shot object detection.
\newblock In \emph{{IEEE} Conference on Computer Vision and Pattern
  Recognition, {CVPR} 2021, virtual, June 19-25, 2021}, pages 13008--13017.
  Computer Vision Foundation / {IEEE}, 2021.

\bibitem[Zhu et~al.(2021{\natexlab{a}})Zhu, Chen, Ahmed, Shen, and
  Savvides]{Zhu21SRR-FSD}
Chenchen Zhu, Fangyi Chen, Uzair Ahmed, Zhiqiang Shen, and Marios Savvides.
\newblock Semantic relation reasoning for shot-stable few-shot object
  detection.
\newblock In \emph{2021 IEEE/CVF Conference on Computer Vision and Pattern
  Recognition (CVPR)}, pages 8778--8787, 2021{\natexlab{a}}.

\bibitem[Zhu et~al.(2021{\natexlab{b}})Zhu, Su, Lu, Li, Wang, and
  Dai]{zhu2021deformable}
Xizhou Zhu, Weijie Su, Lewei Lu, Bin Li, Xiaogang Wang, and Jifeng Dai.
\newblock Deformable {\{}detr{\}}: Deformable transformers for end-to-end
  object detection.
\newblock In \emph{International Conference on Learning Representations},
  2021{\natexlab{b}}.

\end{thebibliography}
}


\end{document}